\documentclass[11pt]{article}

\usepackage[margin=1in]{geometry}
\usepackage[T1]{fontenc}
\usepackage{lmodern}
\usepackage{microtype}
\usepackage{amsmath,amssymb}
\usepackage{booktabs}
\usepackage{tabularx}
\usepackage{xcolor}
\usepackage{graphicx}
\usepackage{tikz}
\usepackage{pgfplots}
\pgfplotsset{compat=1.18}
\usepgfplotslibrary{groupplots}
\usepackage[numbers,sort&compress]{natbib}
\usepackage{xurl}
\usepackage[colorlinks=true,linkcolor=rbBlue,citecolor=rbBlue,urlcolor=rbBlue]{hyperref}
\hypersetup{pdftitle={RehearsalBench: Can AI Predict What Happens When You Change a Company?},
  pdfauthor={Mika Siddiqui and Tharaka Perera}}
\usepackage{enumitem}
\setlist{nosep,leftmargin=*}

\newcommand{\RB}{RehearsalBench}

\definecolor{rbGray}{HTML}{CECECE}
\definecolor{rbPurple}{HTML}{A559AA}
\definecolor{rbTeal}{HTML}{59A89C}
\definecolor{rbGold}{HTML}{F0C571}
\definecolor{rbRed}{HTML}{E02B35}
\definecolor{rbBlue}{HTML}{082A54}
\definecolor{rbTealL}{HTML}{BFE0DA}
\definecolor{rbGoldL}{HTML}{F8E3B6}
\definecolor{rbPurpleL}{HTML}{DCC0DE}
\definecolor{rbRedS}{HTML}{E98A8F}
\definecolor{rbGrayL}{HTML}{E8E8E8}
\definecolor{rbInk}{HTML}{2B4669}

\title{\RB: Can AI Predict What Happens When You Change a Company?}

\author{Mika Siddiqui\\ \small Afterflow Inc.
\and
Tharaka Perera\\ \small Independent contributor}
\date{November 2026}

\begin{document}
\maketitle

\begin{abstract}
Organizations are rushing to put artificial intelligence (AI) into customer service, yet they never see
what would have happened had they chosen differently. A contact center observes only the outcome of the
change it makes. As large language models (LLMs) are increasingly asked to recommend such changes, a
central question goes untested: can they anticipate the effect of an intervention before it is made? We
introduce \RB{}, to our knowledge the first benchmark that evaluates both the operational decisions of LLMs
and their forecasts of counterfactual effects. Models act as managers of simulated bank contact centers,
built and validated on historical call records. They read management reports, choose an operational plan,
and forecast the effects of a common set of interventions, from staffing changes to AI voice agents.
Because every plan is replayed on the same simulated calls and every model forecasts the same
interventions, decisions and forecasts can be compared directly across models.

Across ten closed and open-weight LLMs, the strongest models find plans that beat a budget-limited
optimization search in up to half of their runs, and their forecasts far outperform a no-effect baseline.
Yet a shared blind spot emerges: models from every developer predict the direction of each effect
correctly but overstate its size, most strikingly the repeat contacts generated by an AI voice agent, which
they overestimate roughly five-fold. Planning and forecasting skill do not always coincide, and a model's
price is a poor guide to the value of its decisions. The benchmark's forecast ranking also closely matches
how the same models rank on 33 real organizational changes, a link that appears to reflect, in part, how
well models judge their own uncertainty.

\RB{} offers a practical way to rehearse AI-recommended operational changes before they reach real
customers and staff, and a testbed for building decision-support agents that know not only which way a
change will move outcomes, but by how much.
\end{abstract}

\section{Introduction}
\label{sec:intro}

Organizations increasingly turn to artificial intelligence (AI) to reshape customer service. In July
2025, the Commonwealth Bank of Australia announced that an AI voice bot would replace 45
customer-service roles. The staff union subsequently reported rising call volumes. In August, the bank
reversed its decision, acknowledging that the roles were still needed~\citep{infoage2025cba}. This case
illustrates a general difficulty. Even when a bot resolves some calls, managers must anticipate how much
staff time the remaining calls will require, and they must do so before the change is made.

Moreover, the effect of a change is difficult to measure even afterwards. A contact center, the team that
handles an organization's customer calls, observes the outcome of only the decision it actually makes. If
it moves staff to an evening shift, it can measure how long callers then wait. However, the wait under the
old roster remains a \emph{counterfactual}, an outcome that was never observed. A simulation can test
both rosters against the same calls. Nevertheless, how well such a comparison represents a real center
depends on the simulator's assumptions.

AI models are increasingly evaluated both as forecasters and as agents that run simulated businesses
(Section~\ref{sec:related}). Forecasting benchmarks score predictions of world events. Business
simulations score the outcomes of each agent's own actions. Neither tests whether a model can predict the
effect of an operational change before it is made. Neither compares different models' forecasts of the
same change.

\begin{figure}[t]
\centering
\includegraphics[width=\linewidth]{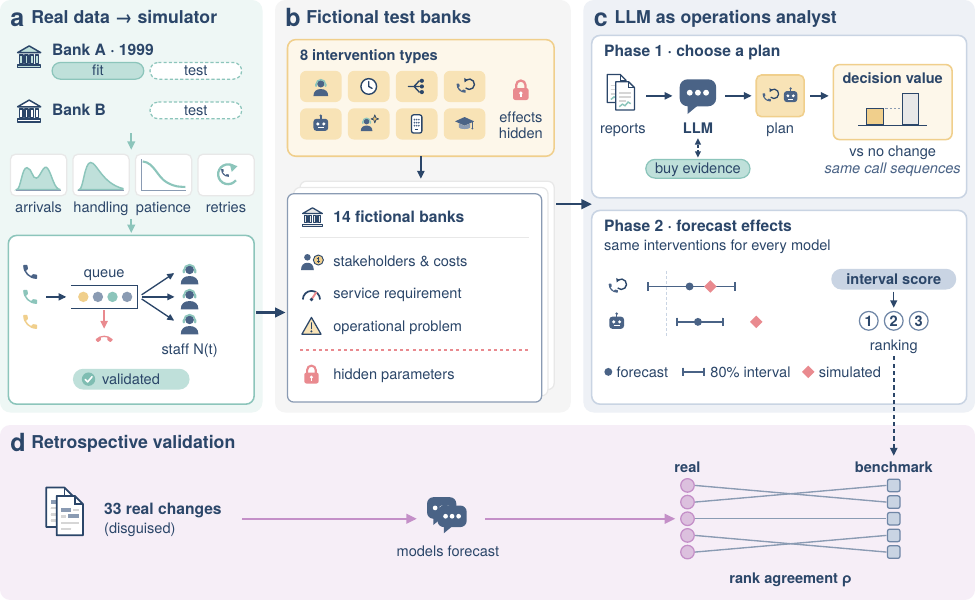}
\caption{Overview of \RB{}. \textbf{(a)} Statistics estimated from Bank~A's records calibrate a
multi-class queue simulator, which is validated on held-out Bank~A records and on Bank~B. \textbf{(b)} The
simulator generates 14 fictional banks, each with stakeholders, costs and an operational problem. Bank
parameters and the effects of the eight intervention types (icons, left to right and top to bottom:
staffing, opening hours, routing, callbacks, AI voice agent, AI assistance for staff, self-service and
coaching) stay hidden from the model. \textbf{(c)} In phase~1, the LLM chooses a plan, scored by its
decision value relative to no change on the same call sequences. In phase~2, every model forecasts the
effects of the same assigned interventions, scored with the interval score. \textbf{(d)} Models also
forecast 33 disguised real organizational changes, and the two forecast rankings are compared. Example
plans, forecasts and rankings are illustrative.}
\label{fig:design}
\end{figure}

We therefore propose \RB{}, a benchmark that asks large language models (LLMs) to rehearse
operational decisions in simulated bank contact centers (Figure~\ref{fig:design}). To the best of our knowledge, \RB{} is the first
benchmark that evaluates both the operational decisions of LLMs and their forecasts of the counterfactual
effects of shared interventions. It is also the first to validate the resulting forecast ranking against
real organizational changes. We examine two questions:
\begin{itemize}
\item \textbf{First, can LLMs choose a valuable intervention and predict its effects?} In \RB{}, each
  model chooses a plan for a simulated center. It then forecasts the effects of a fixed list of
  interventions given to every model. This allows forecasts to be compared even when models recommend
  different plans.
\item \textbf{Second, does the resulting forecast ranking carry over to real organizational changes?} We
  give the models historical cases, using only information available before each change, and rank their
  forecasts against the observed outcomes.
\end{itemize}
The second comparison tests whether forecasting skill transfers across settings. It does not estimate the value
of implementing a recommended plan in a real center.

The main contributions of this paper are as follows:
\begin{enumerate}
\item \emph{Counterfactual benchmark:} We develop a simulator of bank contact centers that is built and
  validated on historical call records. It is combined with explicit, documented assumptions about the
  effects of AI interventions. Because every alternative plan faces the same simulated calls, the
  benchmark measures the value of each plan relative to making no change.
\item \emph{Comparable forecasts:} We separate planning from forecasting. Every model forecasts the same
  assigned interventions with point estimates and prediction intervals, so forecasts can be ranked fairly
  even when the models' own plans differ.
\item \emph{Evaluation of open-weight and closed models:} We evaluate open-weight and closed LLMs on
  simulated contact centers and analyze where their forecasts are incorrect. On average, every model
  predicts the direction of each effect correctly, but all models overstate the size of several. These include the
  increase in repeat contacts under a voice agent, which models overstate approximately five-fold.
\item \emph{Retrospective validation:} We test whether the benchmark's forecast ranking agrees with the
  models' ranking on real organizational changes. The test uses pre-registered criteria and checks for
  prior knowledge of the outcomes.
\end{enumerate}

The remainder of this paper is organized as follows. Section~\ref{sec:related} reviews related work.
Section~\ref{sec:environment} describes the simulation environment, and Section~\ref{sec:benchmark} the
benchmark tasks and scoring. Section~\ref{sec:setup} presents the experimental setup and
Section~\ref{sec:results} the results. Section~\ref{sec:behavior} analyzes
forecast errors and planning behavior. Section~\ref{sec:external} compares the benchmark ranking with
performance on real organizational changes. Section~\ref{sec:conclusion} concludes the
paper and outlines future work.

\section{Related Work}
\label{sec:related}

In this section, we review three lines of work: i) AI forecasting benchmarks and data memorization, ii)
evaluations of AI agents in business and professional tasks, and iii) queueing models of contact centers.

\paragraph{Forecasting benchmarks and data memorization.}
\citet{karger2024forecastbench} propose ForecastBench, which poses questions about future events drawn
from prediction markets, forecasting platforms and public time series. It scores probability forecasts
with the Brier score as the events resolve, using crowd forecasts as a provisional reference for
unresolved market questions. This design prevents models from knowing the answer but requires waiting for
outcomes. Notably, expert forecasters outperform the best LLM.

Retrospective tests can use known outcomes immediately. However, \citet{lopezlira2025memorization} show
that GPT-4o recalls exact economic and financial values from before its knowledge cutoff. Masking fails,
because the model reconstructs the company, quarter and year of an anonymized earnings call from minimal
context. \citet{bordt2024elephants} find that LLMs have memorized popular tabular datasets verbatim and
predict better on these datasets than on novel ones, which the authors attribute to overfitting. We
address this risk with disguised events and recognition checks (Section~\ref{sec:recognition}).

In contrast to \RB{}, forecasting benchmarks such as ForecastBench score forecasts of events that occur
regardless of any decision. They do not score forecasts of what an intervention would change relative to
the counterfactual without it.

\paragraph{Agents in business and professional work.}
\citet{backlund2025vendingbench} propose Vending-Bench, in which an agent manages inventory, orders and
prices for a simulated vending business and is scored on its final net worth. Every model tested has runs
that derail. \citet{chen2026ceobench} propose CEO-Bench, in which an agent runs a fictional
subscription-software company for 500 simulated days and is scored on its final cash. Every model scores
below a rule-based baseline. CEO-Bench also asks for four-week cash forecasts, but these concern each
agent's own company under its own decisions. \citet{park2023generative} place 25 generative agents in a
simulated town and have human evaluators rank the believability of their behavior, without scoring their
decisions against outcomes.

Other benchmarks assess professional work directly. \citet{patwardhan2025gdpval} propose GDPval, in which
occupational experts blindly compare model and human deliverables on tasks drawn from 44 occupations.
\citet{jimenez2024swebench} propose SWE-bench, in which a model edits a Python repository to resolve a real
GitHub issue and is scored by the repository's tests. SWEeper-Bench~\citep{yao2026sweeperbench} gives an
agent only a web application's codebase and an open-ended request. The agent must find the bug through
interactive testing before repairing it, and the best of 15 agents passes 59.0\% of tasks.
$\tau$-bench~\citep{yao2024taubench} has an agent use tools and follow domain policies to serve an
LLM-simulated customer in retail and airline service. Success is judged by the final database state, and
even the strongest agents succeed inconsistently across repeated trials. \citet{kwa2025measuring} at METR
(Model Evaluation and Threat Research) measure how long humans with relevant expertise typically take to
complete tasks that agents complete with 50\% success. They find that this time horizon has doubled
approximately every seven months since 2019.

Nevertheless, all of these benchmarks score the outcomes of each agent's own actions. Because agents act
differently, the benchmarks cannot compare how well different models predict the effect of the same
change. \RB{} measures this directly by having every model forecast the same assigned interventions.

\paragraph{Queueing models of contact centers.}
\citet{gans2003telephone} review call-center operations, covering arrival forecasting, staffing,
abandonment and skills-based routing. They note that the Erlang-C model common in practice ignores customer
impatience. \citet{garnett2002designing} analyze the Erlang-A model, in which callers with exponentially
distributed patience may hang up before being answered. They derive staffing rules of thumb for large
call centers and show that staffing tools that ignore abandonment lead to overstaffing.
\citet{brown2005statistical} use a year of call-by-call records from Bank~A (Section~\ref{sec:records}) to
test these models against observed behavior. They find a strong linear relation between average wait and
the fraction of callers who abandon, and show that Erlang-A gives useful upper bounds on the main
performance measures.

Our simulator extends this line of work. It models time-varying arrivals, staffing and caller patience
(Section~\ref{sec:model}) and draws on the same archive of call records. However, rather than using the
model to design staffing, we use it as a testbed to score AI decisions and forecasts.

\section{Simulation Environment}
\label{sec:environment}

In this section, we describe the environment in which models are evaluated. We first summarize the
statistics estimated from historical call records (Section~\ref{sec:records}) and the contact-center model
built on them (Section~\ref{sec:model}). We then describe how the simulator is validated and how the
validation data relate to the benchmark test set (Section~\ref{sec:validation}). Finally, we describe the
assumed effects of AI interventions (Section~\ref{sec:effects}).

\subsection{Historical Call Records}
\label{sec:records}

Our simulator is built on the call archive of the Service Enterprise Engineering Laboratory (SEELab) at
the Technion -- Israel Institute of Technology~\citep{brown2005statistical}. For each call, the records
give its arrival time, its type and, if it was answered, how long staff spent handling it. Some callers
hang up before being answered, which is known as \emph{abandonment}. For these callers, the records also
give how long they waited. The records also show how many staff were on duty in each interval.
We use records from two banks, which we refer to as \emph{Bank~A} and \emph{Bank~B}. Both banks are
anonymized in the archive. Bank~A's 1999 records are those analyzed by \citet{brown2005statistical}.
Bank~B is a US bank whose records cover the consumer-loan line of its contact center from 2001 to 2003.

From these records, we estimate four statistics for each call type $k$:
\begin{enumerate}
\item \emph{Arrival rates.} The rate at which type-$k$ calls arrive in interval $t$ on weekday type $d$,
\begin{equation}
  \hat{\lambda}_k(d,t) = \frac{1}{|\mathcal{D}_d|\,\tau} \sum_{\ell \in \mathcal{D}_d} M_k(\ell,t),
  \label{eq:arrivals}
\end{equation}
where $\mathcal{D}_d$ denotes the set of recorded days of weekday type $d$, $M_k(\ell,t)$ represents the
number of type-$k$ calls that arrived on day $\ell$ in interval $t$, and $\tau$ corresponds to the
interval length (one hour).
\item \emph{Handling times.} The distribution $\hat{G}_k$ of the time staff spend on answered type-$k$
  calls.
\item \emph{Patience.} The distribution $\hat{F}_k$ of how long type-$k$ callers are willing to wait
  before hanging up. For an abandoned call, the recorded wait equals the caller's patience. For an
  answered call, it is only a lower bound, because the caller would have waited longer. We therefore
  estimate $\hat{F}_k$ with the Kaplan--Meier estimator, which accounts for such censored observations.
\item \emph{Retry and repeat probabilities.} The probability $\hat{p}_k$ that an abandoned caller calls
  again, and the probability $\hat{r}_k$ that an answered call is followed by a repeat contact. Both are
  estimated as the share of callers who call again within 24 hours, using the half of the calls whose
  callers are identified.
\end{enumerate}

\subsection{Contact-Center Model}
\label{sec:model}

We model a contact center as a multi-class queue with time-varying arrivals, impatient callers and a
staffing roster. Calls of type $k$ arrive as a non-homogeneous Poisson process with rate $\lambda_k(t)$.
Each call $i$ has an arrival time $a_i$, a type $k_i$, a handling time $h_i$ and a patience $\theta_i$,
drawn as
\begin{equation}
  h_i \sim G_{k_i}, \qquad \theta_i \sim F_{k_i},
  \label{eq:call}
\end{equation}
where $G_k$ denotes the handling-time distribution of type-$k$ calls and $F_k$ represents their patience
distribution. The roster sets the number of staff on duty, $N(t)$, and a routing policy $\pi$ assigns
waiting calls to available staff. Let $W_i$ denote the time call $i$ would wait before being answered under
$(N,\pi)$. The caller abandons if this wait exceeds their patience,
\begin{equation}
  A_i = \mathbf{1}\{W_i \geq \theta_i\},
  \label{eq:abandon}
\end{equation}
where $\mathbf{1}\{\cdot\}$ is the indicator function. An abandoned caller calls again with probability
$p_k$; otherwise, the contact is \emph{lost}. An answered call leads to a repeat contact with probability
$r_k$.

Over a horizon with $n$ calls, the simulator reports five outcomes: the
abandonment rate $\alpha = \frac{1}{n}\sum_i A_i$, the mean wait $\bar{w} = \frac{1}{n}\sum_i
\min(W_i,\theta_i)$, the average handling time $\bar{h} = \sum_i (1-A_i)h_i \big/ \sum_i (1-A_i)$, the
repeat-contact rate $\gamma$, and the staff cost $C_{\mathrm{staff}} = c_{\mathrm{staff}} \sum_t N(t)\,
\tau$, where $c_{\mathrm{staff}}$ is the hourly cost of one staff member.

We collect the bank's parameters in $\omega = (\{\lambda_k\},\{G_k\},\{F_k\},\{p_k\},\{r_k\})$. A plan $x$
specifies the roster, the routing policy and any interventions. For a random call sequence $\xi$, which fixes
the draws of arrivals, handling times and patience, the simulator returns the outcome vector
\begin{equation}
  \mathbf{Y}(x;\omega,\xi) = \big(\alpha,\ \bar{w},\ \bar{h},\ \gamma,\ C_{\mathrm{staff}}\big).
  \label{eq:outcomes}
\end{equation}
Every plan is replayed on the same call sequences, so differences between plans reflect the plans
themselves rather than random variation. The \emph{effect} of
plan $x$ on outcome $m$ is
\begin{equation}
  \Delta_m(x) = \mathbb{E}_{\xi}\big[\, Y_m(x;\omega,\xi) - Y_m(x_0;\omega,\xi) \,\big],
  \label{eq:effect}
\end{equation}
where $x_0$ denotes current operations and the expectation is taken over call sequences. Models forecast
this effect, together with its variation across call sequences (Section~\ref{sec:tasks}).

\subsection{Simulator Validation and Data Roles}
\label{sec:validation}

We validated the simulator in two settings. In each, we gave it the recorded call arrival times and the
number of staff on duty. It then simulated how the calls were handled, including how long callers waited
and whether they hung up. We compared the simulated outcomes with the records, using criteria set before
each test.
\begin{enumerate}
\item \textbf{Later months at the same bank.} We estimated the statistics in Section~\ref{sec:records}
  from Bank~A's January--June 1999 records. We then compared the simulator's results with its
  July--December records, which we had set aside for testing.
\item \textbf{A different bank.} We tested whether the same simulator could reproduce outcomes at another
  center, using Bank~B's 2001--2003 records from its consumer-loan line.
\end{enumerate}
In both settings, the simulator reproduced the aggregate outcomes in the records. However, two features varied over time in the records: the way
calls were assigned to staff (\emph{routing}) and the level of service callers received. Rather than
fixing either at a single value, we sample both from ranges, so that the simulated banks reflect this
variation.

Table~\ref{tab:dataroles} summarizes how the data sets relate to one another. The validation data test the
simulator, whereas the benchmark test set tests the models. The simulator connects the two. Once validated,
it generates each test bank by drawing parameters $\omega$ around the estimated statistics. Every test bank's history is built from Bank~A's fit, with bank-level
values drawn within a declared envelope. For example, handling times are 1.00--1.36 times and caller
patience 1.0--2.4 times Bank~A's, and call volumes range from 4,700 to 17,300 calls a week. The test banks are
therefore grounded in the records, but none is a copy of Bank~A or Bank~B. The real organizational changes in
Section~\ref{sec:external} form a separate test that does not use the simulator.

\begin{table}[htbp]
\centering
\small
\caption{Data sets and their roles.}
\label{tab:dataroles}
\begin{tabularx}{\linewidth}{@{}lXl@{}}
\toprule
Data & Role & Used in \\
\midrule
Bank~A, Jan--Jun 1999 & Estimate call statistics (calibration) & Section~\ref{sec:records} \\
Bank~A, Jul--Dec 1999 & Held-out validation: same bank, later period & Section~\ref{sec:validation} \\
Bank~B, consumer loans, 2001--2003 & External validation: different bank & Section~\ref{sec:validation} \\
14 fictional banks & Benchmark test set, generated by the validated simulator & Sections~\ref{sec:results}--\ref{sec:behavior} \\
33 real organizational changes & Retrospective test of the forecast ranking & Section~\ref{sec:external} \\
\bottomrule
\end{tabularx}
\end{table}

\subsection{Assumed Intervention Effects}
\label{sec:effects}

An \emph{intervention} is a single operational change, such as adding evening staff or deploying an AI
voice agent. A \emph{plan} combines one or more interventions, or keeps current operations.
Table~\ref{tab:interventions} lists the eight types of intervention available to models and the part of
the model in Section~\ref{sec:model} that each one changes.

\begin{table}[htbp]
\centering
\small
\caption{Intervention types and the model components they change.}
\label{tab:interventions}
\begin{tabularx}{\linewidth}{@{}lXl@{}}
\toprule
Intervention & What it does & Changes \\
\midrule
Staffing & Changes how many staff are on duty in each interval & $N(t)$ \\
Opening hours & Extends or shortens the hours in which calls are answered & $N(t)$ \\
Routing & Changes how waiting calls are assigned to staff & $\pi$ \\
Callbacks & Offers waiting callers a call back instead of holding & $F_k$, $\lambda_k(t)$ \\
AI voice agent & Resolves some calls without staff & $\lambda_k(t)$, $G_k$, $r_k$ \\
AI assistance for staff & Helps staff handle calls faster & $G_k$ \\
Self-service & Lets customers resolve requests online instead of calling & $\lambda_k(t)$ \\
Coaching & Trains staff to resolve calls faster or at the first attempt & $G_k$, $r_k$ \\
\bottomrule
\end{tabularx}
\end{table}
Staffing, opening hours and routing change only how existing calls are served, so their effects follow
from the model in Section~\ref{sec:model}. The other interventions also change the calls themselves. The
records describe ordinary call handling, so simulating these interventions requires assumptions about their
effects. For example, what share of calls could an AI voice agent resolve without staff? Each
intervention $j$ modifies the model through a small set of effect parameters $\psi_j$. For example, a voice
agent resolves a share $c_k$ of type-$k$ calls without staff and changes the repeat-contact probability,
\begin{equation}
  \lambda_k(t) \rightarrow (1-c_k)\,\lambda_k(t), \qquad r_k \rightarrow r_k + \Delta r_{\mathrm{va}},
  \label{eq:voice}
\end{equation}
where $c_k$ denotes the containment share and $\Delta r_{\mathrm{va}}$ represents the change in the
repeat-contact probability. The voice agent also changes the handling times of the calls left to staff. A call that the voice agent does not resolve reaches the queue later, and its handling time is longer by 0--90~s. Containment does not depend on the call's handling time. Similarly, AI assistance for staff shortens handling times,
$h_i \rightarrow (1-\delta)\,h_i$, where $\delta$ is the fractional time saving. 

Self-service, callbacks
and coaching act on arrival rates, patience and handling times. Self-service removes a share of the arriving calls of the targeted types (20--36\% for simple types). Callbacks let callers whose expected wait exceeds a threshold keep their place in the queue without abandoning. Coaching moves each coached staff member's handling time toward that of the fastest fifth of staff
(9--25\% of the gap) and takes the coaching hours off the phones.

We draw the effect parameters from documented ranges. For example, voice-agent containment is drawn from
14--36\%, following a 2024 Gartner customer survey~\citep{gartner2024selfservice}, and AI assistance
shortens handling times for existing staff by 0--13\%, following~\citet{brynjolfsson2025generative}.
Parameters with no published measurement, such as the share of callers who agree to use the voice agent,
are assumed and given wide ranges. For each bank, the drawn values,
together with $\omega$, are kept hidden from the model being evaluated. Figure~\ref{fig:design}
illustrates how the historical records and the assumed effects combine to support the two evaluations.

\section{Benchmark Design}
\label{sec:benchmark}

In this section, we describe how test banks are constructed and selected
(Section~\ref{sec:banks}), the two tasks given to each model (Section~\ref{sec:tasks}), and how plans and forecasts are scored
(Section~\ref{sec:scoring}).

\subsection{Simulated Banks and Selection}
\label{sec:banks}

Each benchmark instance is a fictional bank (for example, bank B04) whose customer contact center is
simulated with the model of Section~\ref{sec:model}. The bank provides the organizational context for the
decision: stakeholders such as Finance, their goals, the monetary cost they assign to repeat
and lost contacts, a minimum service requirement, and an operational problem to address. The contact
center is the operation that the model changes. We therefore refer to each instance by its bank and to the operation being changed as its contact
center.

For each bank, we generate 26 weeks of management reports, which the model receives together with the
roster, a cost sheet and stakeholder statements. A search with a fixed budget
of 540 plan evaluations then looks for the most valuable plan it can find, using separate call sequences
that are never used for scoring. We call the result
the \emph{best-known plan}. It serves as a reference rather than an optimum, and models frequently
outperform it (Section~\ref{sec:results}).

Before a candidate bank is used, it must pass three screening checks. These checks confirm that i) current
operations are feasible, ii) the hidden parameters cannot be read from the material given to the model,
and iii) the best-known plan gains over both making no change and a generic plan chosen without the bank's
data. We record both accepted and rejected candidates.

Candidate banks were generated in two batches. The first batch contained 24 candidates, three for each of
eight intervention types, of which 21 passed screening. A pre-specified balance rule required the best-known plans
to cover a mix of intervention types. However, staffing changes dominated the best-known plans, so the
batch failed this rule and its two fallback rules. The final fallback therefore selected eight banks that
passed screening: for each of the eight intervention types, taken in a fixed order, it selected the first
accepted bank whose specification targeted that type. A second batch of 16 candidates added six banks, bringing the total to
14. Two of these six are banks where no available plan reliably outperforms making no change. They test
whether models recognize when keeping current operations is the best choice. Three others impose a hiring
freeze.

\subsection{Model Tasks}
\label{sec:tasks}

A \emph{run} is one attempt by one model on one bank. Each run consists of two phases, carried out in an
environment without network access.
\begin{enumerate}
\item \textbf{Phase 1: choose a plan.} The model reads the bank's management reports, roster, cost sheet
  and stakeholder statements. It may purchase evidence, such as a short pilot, staff-level records or
  vendor references, at a cost charged to its plan. A pilot is a simulated trial of one intervention,
  lasting two, four or eight weeks, on about a third of the staff or calls, run on call sequences separate
  from those used for scoring. It costs \$15,000 per two weeks, and the model sees only its noisy
  readout. It then ranks the bottlenecks it identifies, chooses a
  plan and forecasts the plan's outcomes. The plan may keep current operations or combine any of the
  interventions in Table~\ref{tab:interventions}.
  We then lock the submission.
\item \textbf{Phase 2: forecast assigned interventions.} In a fresh conversation, every model receives the
  same list of specific interventions for the bank, such as deploying an AI voice agent or adding evening
  staff. We call these the \emph{assigned interventions}. They are generated by a fixed rule, the same for every bank: a cut of about two staff members, an AI voice agent, callbacks offered when the expected wait exceeds
  five minutes, AI assistance for all staff, self-service content for the largest simple call type, two
  hours of coaching per staff member per month and, in banks whose staff form a single team, two staff
  members reserved for specialist call types. Banks in the second batch also get a roster reshape that
  moves two staff members. For each
  intervention and each of the five outcomes in Eq.~\eqref{eq:outcomes}, the model gives a point estimate
  of the effect $\Delta_m$ in Eq.~\eqref{eq:effect} and an 80\% prediction interval, a range that it
  expects to contain the simulated effect four times in five (models also forecast outcome levels, but only the stated effects are scored).
\end{enumerate}
Phase~1 is limited to 400,000 generated tokens, 60 minutes and 200 tool calls per run, and phase~2 to
300,000 generated tokens, 30 minutes and 100 tool calls.

We report two results for each model: the value of its own plan from phase~1 (decision
value) and the accuracy of its assigned forecasts from phase~2 (forecast improvement). Because every model
forecasts the same interventions, the phase~2 results yield the forecast ranking that we compare with real
organizational changes in Section~\ref{sec:external}.

\subsection{Scoring}
\label{sec:scoring}

\paragraph{Decision value.}
For each bank, the evaluator simulates four alternatives: current operations $x_0$ (no change), the
model's plan, the assigned interventions and the best-known plan. Each alternative is simulated in three
versions of the bank: as described to the model, under adverse conditions and under favorable conditions. The versions differ only in assumed parameters, such as intervention effects, caller patience and the stakeholders' valuations. The described version draws each from its published range, the adverse version from the half less favorable to automation and AI, and the favorable version from the other half, with no cost for lost contacts. The mechanics estimated from the records are the same in all three. In each version, every alternative is replayed on the same
$R$ randomly generated call sequences $\xi_1,\dots,\xi_R$, each covering a 13-week horizon.

The utility of a plan $x$ is
\begin{equation}
  U(x) = -\big(C_{\mathrm{staff}}(x) + C_{\mathrm{int}}(x)\big)
         - v_{\mathrm{repeat}}\,N_{\mathrm{repeat}}(x) - v_{\mathrm{lost}}\,N_{\mathrm{lost}}(x),
  \label{eq:utility}
\end{equation}
where $C_{\mathrm{staff}}$ denotes the staff cost, $C_{\mathrm{int}}$ represents the cost of the plan's
interventions and purchased evidence, with one-off costs spread over 24 months, $N_{\mathrm{repeat}}$ and $N_{\mathrm{lost}}$ correspond to the
numbers of repeat contacts and lost contacts, and $v_{\mathrm{repeat}}$ and $v_{\mathrm{lost}}$ are the
dollar costs that the bank's stakeholders assign to each. The \emph{decision value} of plan $x$ is its gain
over making no change, expressed as a percentage of the total no-change cost:
\begin{equation}
  \mathrm{DV}(x) = 100 \times \frac{\bar{U}(x) - \bar{U}(x_0)}{-\bar{U}(x_0)},
  \label{eq:dv}
\end{equation}
where $\bar{U}$ denotes utility averaged over the three versions and $R$ call sequences, and $-\bar{U}(x_0)$ is
the total cost of making no change, including the costs of repeat and lost contacts. A positive decision value therefore means that the plan
outperforms making no change. A plan can gain value by reducing repeat or lost contacts as well as direct
costs. Evidence purchases are charged to the plan. If a plan breaches the bank's minimum service requirement in any version,
it earns no credit: it is scored as making no change, so its decision value is zero.

\paragraph{Forecast score.}
Decisions depend on how much an intervention changes outcomes, so we score forecasts of effects (Eq.~\eqref{eq:effect})
rather than of outcome levels.

We score each 80\% prediction interval with the interval score~\citep{gneiting2007scoring},
\begin{equation}
  \mathrm{IS}(l,u;y) = (u-l) + \frac{2}{\beta}(l-y)\,\mathbf{1}\{y<l\} + \frac{2}{\beta}(y-u)\,\mathbf{1}\{y>u\},
  \label{eq:is}
\end{equation}
where $l$ and $u$ denote the lower and upper bounds of the interval, $y$ represents the simulated effect,
and $\beta=0.2$ corresponds to the share of outcomes an 80\% interval is allowed to miss. The first term
penalizes wide intervals, and the other two penalize intervals that miss the simulated effect, in
proportion to the size of the miss. We compute the score against the effect in each of the $R$ call
sequences of the version described to the model, that is, the paired difference inside the expectation in
Eq.~\eqref{eq:effect}, and average over the sequences. To make outcomes on different scales comparable, we divide each score
by $\sigma_m$, the standard deviation of outcome $m$ under no change across call sequences. We then average the
scaled scores over outcomes, interventions and banks to obtain a model's forecast score
$S_{\mathrm{model}}$.

The \emph{forecast improvement} compares this score with that of a baseline that always predicts zero
effect with a zero-width interval,
\begin{equation}
  \mathrm{FI} = 1 - \frac{S_{\mathrm{model}}}{S_{0}},
  \label{eq:fi}
\end{equation}
where $S_{0}=10.19$ denotes the baseline's score in this evaluation. A forecast improvement of zero
matches the baseline, and one corresponds to perfect forecasts. We report it as a percentage.

\paragraph{Uncertainty in comparisons.}
A model's score depends on which banks are in the benchmark and on how its runs vary on the same bank. To
capture both sources of uncertainty, we compute 95\% bootstrap intervals from 10,000 resamples. Each
resample first draws banks with replacement and then draws runs within each sampled bank. We use the same
draws for all models, so that every model is compared on the same resampled benchmark. We consider two
models that are next to each other in a ranking to be distinguishable only when the interval for the
difference between their scores excludes zero. We do not adjust these intervals for multiple comparisons,
so individual distinctions should be read as indicative. The comparison with real organizational changes in
Section~\ref{sec:external} has a different source of uncertainty, namely which real changes are included.
There, we therefore resample changes instead of banks.

\paragraph{Protocol.}
Before running the models, we fixed the bank specifications, call sequences, evaluation software and scoring
rules, and recorded their SHA-256 digests, cryptographic fingerprints that change if any file changes
(listed in the public repository). We recorded all scoring amendments
for the main evaluation before scoring it. Claude Opus~5.5 was used as a coding assistant in developing
the benchmark software and was not among the evaluated models. Because three evaluated models
(Claude Sonnet~5.5, Claude Fable~5.1 and Claude Haiku~5.5) share its developer, we disclose this use.

\paragraph{Remark (relation to reinforcement learning).}
The simulator plays a role similar to an environment in reinforcement learning (RL). The bank's parameters
$(\omega,\psi)$ form a hidden state, and the management reports are observations. Evidence purchases are
information-gathering actions with a cost, and the submitted plan is a final action whose reward is its
decision value. A short pilot is the closest analog of exploration: the model pays to observe a brief,
noisy run of an intervention in the simulator. However, unlike
in standard RL, each run is a single episode, and the model receives no reward until scoring.
\RB{} therefore tests planning under partial observability rather than learning by trial and error.
Nevertheless, the same environment could be used to train agents with RL in future work.

\section{Experimental Setup}
\label{sec:setup}

In this section, we describe the models evaluated, how they were accessed and which runs count toward the
results.

We evaluated ten LLMs from six developers (Table~\ref{tab:models}). Two of them, DeepSeek V4.1 Flash and
Mistral Small 2603, have openly available weights. The other eight, including Qwen3.8 Flash, are available
only through APIs. Each model made two runs on each of the 14
banks, giving 280 runs in total, carried out on 30--31 October 2026. Each run took 40--90 minutes. The
evaluator simulated each alternative on $R=200$ call sequences in each version of a bank. Access to the
models through their application programming interfaces (APIs) cost \$1,202 for this evaluation.
Every model ran at its highest available reasoning setting.

We rank the nine models that submitted a complete plan and forecasts in at least half of their runs, and
report a \emph{partial result} for Mistral Small 2603 separately. Four further models were
evaluated for the comparison with real organizational changes, and we introduce them in
Section~\ref{sec:external}.

\begin{table}[t]
\centering
\small
\caption{Models evaluated and their access routes (28 runs per model). The rest of the
paper refers to models by their short names. OpenRouter is a third-party service that routes requests to
model providers; ``own key'' means that requests were billed to our own account with the provider. Claude
Fable~5.1 used OpenRouter on five banks and its developer's API on nine.}
\label{tab:models}
\begin{tabular}{llll}
\toprule
Model & Short name & Developer & Access route \\
\midrule
GPT-6 Astra & Astra & OpenAI & Developer API \\
GPT-6.1 Sol & Sol & OpenAI & Developer API \\
GPT-6 Luna & Luna & OpenAI & OpenRouter \\
Claude Sonnet~5.5 & Sonnet & Anthropic & OpenRouter \\
Claude Fable~5.1 & Fable & Anthropic & Both \\
Claude Haiku~5.5 & Haiku & Anthropic & Developer API \\
Gemini~3.8 Flash & Gemini Flash & Google & Developer API \\
DeepSeek V4.1 Flash & DeepSeek & DeepSeek & OpenRouter \\
Qwen3.8 Flash & Qwen Flash & Alibaba & OpenRouter (own key) \\
Mistral Small 2603 & Mistral Small & Mistral & OpenRouter \\
\bottomrule
\end{tabular}
\end{table}

\section{Results}
\label{sec:results}

In this section, we first illustrate a single run (Section~\ref{sec:b04}). We then report decision value
and forecast performance for all models (Section~\ref{sec:leaderboard}) and relate them to the cost of
each model (Section~\ref{sec:cost}).

\subsection{Illustrative Run: A Roster Change at Bank B04}
\label{sec:b04}

To illustrate what a run involves, we describe one of Astra's runs. At bank B04, the Finance team
requested a 10\% cut in payroll. Astra found in the management reports that 15,957 weekday calls between
21:00 and midnight had reached an after-hours announcement because no staff were on duty. It proposed moving existing staff hours to evenings and to morning
peaks, rather than adding staff. Astra declined to promise the payroll cut, citing the bank's minimum
service requirement and the cost that Finance assigns to unanswered calls.

The plan improved utility over the 13-week horizon by \$156,683 relative to making no change and by \$17,131 over the best-known plan. Its forecasts were also close to the simulated
outcomes (Table~\ref{tab:b04}). For example, Astra forecast an abandonment rate of 11.0\%, against 10.6\% in
the simulation, and its staff-cost forecast was within \$215 of the simulated cost. This case highlights
the kind of reasoning the benchmark rewards: the model identified a mismatch between when calls arrive and
when staff are on duty, and it weighed a stakeholder's request against the bank's service requirement.

\begin{table}[htbp]
\centering
\small
\caption{Astra's forecasts for its chosen plan at bank B04, compared with the simulated outcomes.
Parentheses give the model's 80\% prediction intervals where it reported them.}
\label{tab:b04}
\begin{tabular}{lrr}
\toprule
Outcome & Forecast & Simulated outcome \\
\midrule
Abandonment rate (\%) & 11.0 (9.8--12.4) & 10.6 \\
Mean wait (seconds) & 40 (33--48) & 38 \\
Staff cost (USD) & 306,185 & 305,970 \\
\bottomrule
\end{tabular}
\end{table}

\subsection{Decision Value and Forecast Performance}
\label{sec:leaderboard}

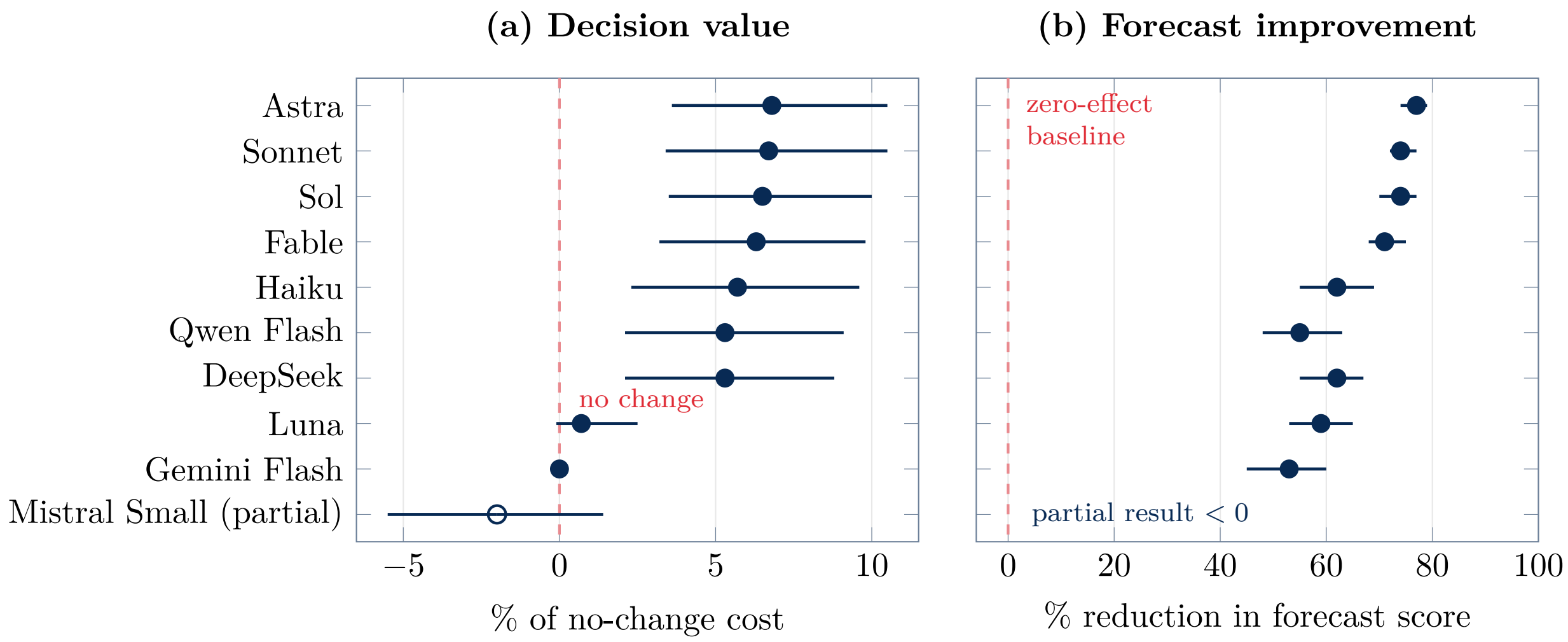
\begin{figure}[t]
\centering
\begin{tikzpicture}
\begin{groupplot}[
  group style={group size=2 by 1, horizontal sep=0.6cm, yticklabels at=edge left},
  width=0.45\linewidth, height=6.4cm, y dir=reverse, ymin=0.4, ymax=10.6,
  ytick={1,...,10}, yticklabels={{Astra},{Sonnet},{Sol},{Fable},{Haiku},{Qwen Flash},{DeepSeek},{Luna},{Gemini Flash},{Mistral Small (partial)}},
  tick label style={font=\small}, label style={font=\small}, title style={font=\small\bfseries},
  xmajorgrids, grid style={rbGrayL}, axis line style={rbBlue!60}, tick style={rbBlue!60},
  error bars/error bar style={rbBlue, line width=0.9pt}, error bars/error mark=none]
\nextgroupplot[title={(a) Decision value}, xlabel={\% of no-change cost}, xmin=-6.5, xmax=11.5]
\draw[rbRedS, dashed, thick] (axis cs:0,0.4) -- (axis cs:0,10.6);
\node[anchor=west, font=\scriptsize, text=rbRed] at (axis cs:0.2,7.5) {no change};
\addplot[only marks, mark=*, mark size=2.6pt, color=rbBlue,
  error bars/.cd, x dir=both, x explicit] coordinates {(6.8,1) += (3.70,0) -= (3.20,0) (6.7,2) += (3.80,0) -= (3.30,0) (6.5,3) += (3.50,0) -= (3.00,0) (6.3,4) += (3.50,0) -= (3.10,0) (5.7,5) += (3.90,0) -= (3.40,0) (5.3,6) += (3.80,0) -= (3.20,0) (5.3,7) += (3.50,0) -= (3.20,0) (0.7,8) += (1.80,0) -= (0.80,0) (0.0,9) += (0.09,0) -= (0.04,0)};
\addplot[only marks, mark=o, mark size=2.6pt, color=rbBlue, thick,
  error bars/.cd, x dir=both, x explicit] coordinates {(-2.0,10) += (3.4,0) -= (3.5,0)};
\nextgroupplot[title={(b) Forecast improvement}, xlabel={\% reduction in forecast score}, xmin=-6, xmax=100]
\draw[rbRedS, dashed, thick] (axis cs:0,0.4) -- (axis cs:0,10.6);
\node[anchor=north west, font=\scriptsize, text=rbRed, align=left] at (axis cs:1,0.45) {zero-effect\\baseline};
\addplot[only marks, mark=*, mark size=2.6pt, color=rbBlue,
  error bars/.cd, x dir=both, x explicit] coordinates {(77,1) += (2.00,0) -= (3.00,0) (74,2) += (3.00,0) -= (2.00,0) (74,3) += (3.00,0) -= (4.00,0) (71,4) += (4.00,0) -= (3.00,0) (62,5) += (7.00,0) -= (7.00,0) (55,6) += (8.00,0) -= (7.00,0) (62,7) += (5.00,0) -= (7.00,0) (59,8) += (6.00,0) -= (6.00,0) (53,9) += (7.00,0) -= (8.00,0)};
\node[anchor=west, font=\scriptsize, text=rbBlue] at (axis cs:2,10) {partial result $<0$};
\end{groupplot}
\end{tikzpicture}
\caption{Main results for the nine ranked models and the partial result for Mistral Small, ordered by
decision value. Dots give means and bars give 95\%
bootstrap intervals. (a) Decision value (Eq.~\eqref{eq:dv}); the open circle marks the partial result for
Mistral Small. (b) Forecast improvement over the zero-effect baseline (Eq.~\eqref{eq:fi}); Mistral Small's
partial result lies below the baseline. API costs per run are shown in Figure~\ref{fig:cost}.}
\label{fig:leaderboard}
\end{figure}

Figure~\ref{fig:leaderboard} shows both results for each model. Seven ranked models chose plans worth
5--7\% of the no-change cost on average, with overlapping bootstrap intervals. Astra had the highest point
estimate (6.8\%), followed closely by Sonnet (6.7\%) and Sol (6.5\%). Among models next to each other in
the ranking, the bootstrap distinguished DeepSeek (5.3\%) from Luna (0.7\%). The ranked models
therefore fall into two tiers: seven models whose plans add similar value, and Luna and Gemini Flash, whose
plans add little. As Section~\ref{sec:behavior} shows, this gap is attributable mainly to plans that left
operations unchanged (Gemini Flash) or cost more than making no change (Luna).

All nine ranked models improved on the zero-effect baseline, with forecast improvements of 53--77\%. Astra
had the best forecasts (77\%), and the bootstrap distinguished it from the next model, Sonnet (74\%). It
also distinguished Fable (71\%) from Haiku (62\%). Notably, planning and forecasting skill do not always
coincide. DeepSeek and Qwen Flash chose plans of the same value (5.3\%), but DeepSeek had the higher forecast improvement
(62\% versus 55\%). Conversely, Luna's plans added little value, yet its forecasts (59\%) were comparable to
Haiku's (62\%).

\paragraph{Remark (comparison with the best-known plan).}
The best-known plan is a reference rather than an upper bound. Although the search evaluates plans
directly in the simulator, its budget is fixed, so it cannot explore every combination of interventions.
In 50\% of its runs, Astra's plan had a higher mean simulated value than the best-known plan for that bank.
Sonnet, Sol and Haiku did so in 39--43\% of runs. This suggests that the strongest models can find plans
that a budget-limited search misses, possibly through the many fine-grained roster changes they make
(Section~\ref{sec:behavior}).

\subsection{Model Cost}
\label{sec:cost}

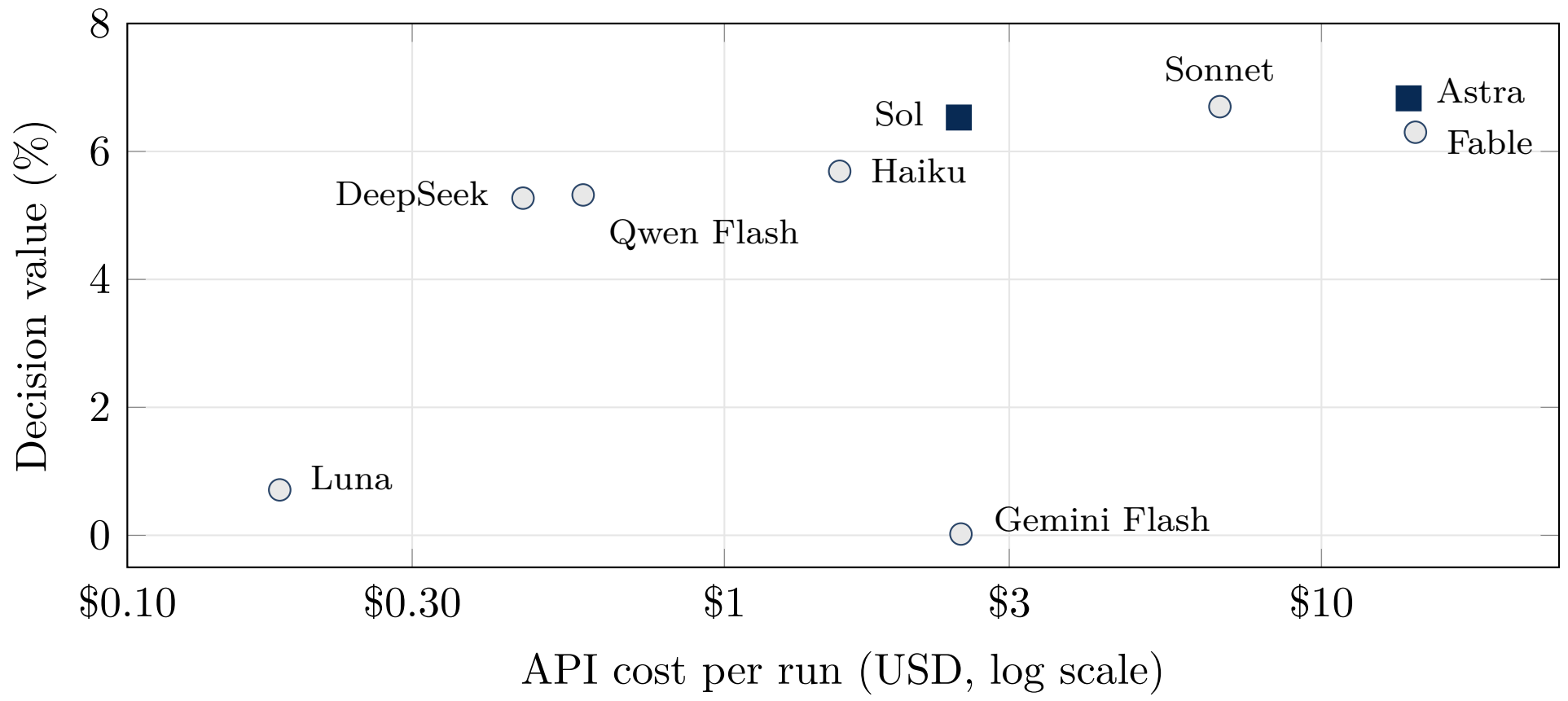
\begin{figure}[htbp]
\centering
\begin{tikzpicture}
\begin{axis}[
  width=0.8\linewidth, height=6cm,
  xmode=log, xmin=0.1, xmax=25, ymin=-0.5, ymax=8,
  xlabel={API cost per run (USD, log scale)}, ylabel={Decision value (\%)},
  xtick={0.1,0.3,1,3,10}, xticklabels={\$0.10,\$0.30,\$1,\$3,\$10},
  grid=major, grid style={gray!20}, tick label style={font=\small}, label style={font=\small}]
\addplot[only marks, mark=*, mark size=2.5pt, fill=rbGrayL, draw=rbInk] coordinates {
  (6.76,6.70) (14.37,6.30) (1.56,5.69) (0.58,5.32) (0.46,5.27) (0.18,0.71) (2.49,0.02)};
\addplot[only marks, mark=square*, mark size=2.8pt, color=rbBlue] coordinates {(14.00,6.83) (2.47,6.53)};
\node[anchor=west, font=\scriptsize] at (axis cs:14.6,6.95) {Astra};
\node[anchor=south, font=\scriptsize] at (axis cs:6.76,6.85) {Sonnet};
\node[anchor=east, font=\scriptsize] at (axis cs:2.3,6.6) {Sol};
\node[anchor=west, font=\scriptsize] at (axis cs:15.2,6.15) {Fable};
\node[anchor=west, font=\scriptsize] at (axis cs:1.65,5.7) {Haiku};
\node[anchor=north west, font=\scriptsize] at (axis cs:0.6,5.2) {Qwen Flash};
\node[anchor=east, font=\scriptsize] at (axis cs:0.43,5.3) {DeepSeek};
\node[anchor=west, font=\scriptsize] at (axis cs:0.19,0.9) {Luna};
\node[anchor=west, font=\scriptsize] at (axis cs:2.65,0.25) {Gemini Flash};
\end{axis}
\end{tikzpicture}
\caption{Mean decision value against API cost per run for the nine ranked models. Squares mark Astra and
Sol, which reach similar value at a six-fold difference in cost.}
\label{fig:cost}
\end{figure}

Figure~\ref{fig:cost} depicts each model's mean decision value against its API cost per run. Decision
value rises only modestly with cost. Sol reached nearly the same value as Astra (6.5\% versus 6.8\%) at
about one-sixth of the cost (\$2.47 versus \$14.00 per run). DeepSeek reached 5.3\%, or about 77\% of
Astra's value, at \$0.46 per run, about one-thirtieth of the cost. In contrast, a higher cost did not
guarantee value. Fable was the most expensive model (\$14.37 per run) but ranked fourth, and Gemini Flash
cost about as much per run as Sol (\$2.49) but added almost no value.
This weak relationship stems from planning behavior rather than price: the two models whose plans added
little value differ 14-fold in cost, and their low value reflects costly or unchanged plans
(Section~\ref{sec:behavior}). Consequently, for this task, a model's cost per run is a poor guide to the
value of its plans.

\section{Forecast Errors and Planning Behavior}
\label{sec:behavior}

In this section, we analyze where models' forecasts are incorrect and how their plans differ, in order to
explain the results in Section~\ref{sec:results}.

\paragraph{Direction of effects.}
On average across runs, every model forecast the correct direction for each assigned effect. This includes
effects whose direction is not obvious. In the simulator, the assigned routing and coaching changes
increased abandonment: coaching takes the coached staff off the phones for their coaching hours, and the routing change reserves
two staff members for specialist call types, leaving fewer agents for the other calls. Similarly, the voice
agent increased the average handling time of the calls left to staff.

\begin{figure}[htbp]
\centering
\begin{tikzpicture}[x=1cm,y=1cm,font=\small]
  \foreach \y/\name/\metric in {0/AI assistance for staff/handling time (s), -1.4/Callbacks/mean wait (s), -2.8/Voice agent/repeat contacts (pp)} {
    \node[anchor=west, font=\small\bfseries] at (-5.2,\y+0.12) {\name};
    \node[anchor=west, font=\scriptsize, text=black!70] at (-5.2,\y-0.25) {change in \metric};
    \draw[gray!40] (0,\y) -- (8,\y);
  }
  \foreach \x/\l in {0/$-14$, 4/$-7$, 8/0} {\draw[gray] (\x,-0.06) -- (\x,0.06); \node[font=\scriptsize, text=black!70] at (\x,-0.32) {\l};}
  \foreach \x/\l in {0/$-6$, 4/$-3$, 8/0} {\draw[gray] (\x,-1.46) -- (\x,-1.34); \node[font=\scriptsize, text=black!70] at (\x,-1.72) {\l};}
  \foreach \x/\l in {0/0, 4/0.3, 8/0.6} {\draw[gray] (\x,-2.86) -- (\x,-2.74); \node[font=\scriptsize, text=black!70] at (\x,-3.12) {\l};}
  \filldraw[fill=rbTealL, draw=rbTeal, rounded corners=1pt] (0.800,-0.12) rectangle (2.857,0.12);
  \filldraw[fill=rbTealL, draw=rbTeal, rounded corners=1pt] (0.800,-1.52) rectangle (6.533,-1.28);
  \filldraw[fill=rbTealL, draw=rbTeal, rounded corners=1pt] (6.133,-2.92) rectangle (7.733,-2.68);
  \fill[rbRed!80] (3.086,0) circle (0.11);
  \fill[rbRed!80] (7.733,-1.4) circle (0.11);
  \fill[rbRed!80] (1.333,-2.8) circle (0.11);
  \filldraw[fill=rbTealL, draw=rbTeal] (0,0.62) rectangle (0.5,0.82); \node[anchor=west, font=\scriptsize] at (0.6,0.72) {range of model means};
  \fill[rbRed!80] (5.0,0.72) circle (0.11); \node[anchor=west, font=\scriptsize] at (5.2,0.72) {simulated effect};
\end{tikzpicture}
\caption{Three effects whose size all models overstated. For each intervention, the band spans the models'
mean point forecasts, and the dot marks the simulated effect, which lies outside the band in all
three cases. Changes in repeat contacts are in percentage points (pp).}
\label{fig:missed}
\end{figure}
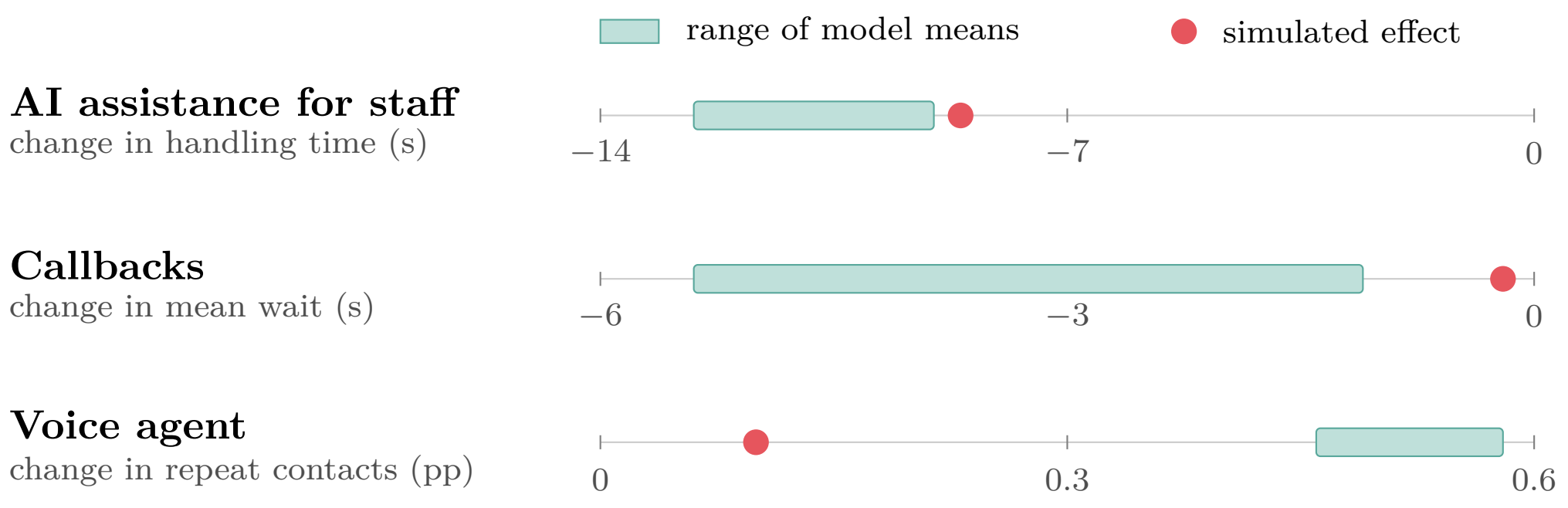

\paragraph{Size of effects.}
However, models from all developers overstated the size of the same three effects
(Figure~\ref{fig:missed}). They predicted that AI assistance would shorten handling time by 9.0--12.6
seconds, against 8.6 seconds in the simulator, an overstatement of 5--47\%. They expected callbacks to
reduce waiting by 1.1--5.4 seconds, against 0.2 seconds, which overstates the effect between 5.5- and
27-fold. Most notably, they forecast that a voice agent would increase repeat contacts by 0.46--0.58
percentage points, against 0.10, approximately five times the simulated effect. Because these errors are
shared across developers, they are unlikely to reflect the weakness of any single model. One possible
explanation is that the vendor materials available to every model overstate these effects. Every bank's stakeholder documents include a voice-agent vendor's note claiming that up to a stated share of calls is fully handled without an agent and that the remaining calls are 30\% shorter. Models can also request vendor reference figures for
any intervention, including AI assistance and callbacks, and these overstate the true effect by an
undisclosed factor of up to about six. We have
not yet tested this, for example by removing those materials. It is worth mentioning that this explanation
fits the overstated benefits of AI assistance and callbacks better than the overstated increase in repeat
contacts, which is a cost of the voice agent.

\paragraph{Keeping current operations.}
No model explicitly recommended keeping current operations, even on the two banks where no available plan
reliably outperformed making no change. Gemini Flash nevertheless submitted plans that left operations
unchanged in 75\% of its runs, including three of its four runs on those two banks. However, most of its
unchanged plans were at banks where an intervention would have added value. This explains its decision
value of 0.0\% (Figure~\ref{fig:leaderboard}). The model rarely acted, and it did not distinguish banks
where acting pays from banks where it does not.

\paragraph{Interval coverage.}
An 80\% interval is well calibrated if it contains the simulated effect about four times in five. Coverage was
80--100\% for the leading forecasters, so their intervals were, if anything, wider
than necessary. In contrast, Haiku's and Gemini Flash's intervals often covered only 11--50\% of the
simulated effects, so they were too narrow. This overconfidence
contributes to their lower forecast improvement (62\% and 53\%), because the interval score penalizes
each miss in proportion to its size (Eq.~\eqref{eq:is}).

\paragraph{Plan composition.}
Staffing changes appeared in 93--100\% of plans for every ranked model except Gemini Flash. This is
consistent with staffing dominating the best-known plans (Section~\ref{sec:banks}). Models differed most
in how finely they adjusted the roster. Astra averaged 46 individual changes per plan, whereas Luna averaged five. Moreover, Luna's plans cost more than making no
change in 43\% of its runs, which explains its low decision value.

The models also differed in their use of
automation. Astra never chose a voice agent, whereas DeepSeek did so in 36\% of its plans and Mistral
Small, with a partial result, in 86\%.

\paragraph{Evidence and forecasts of own plans.}
Astra and Sol purchased about three items of evidence per run, whereas Haiku bought almost none. Haiku's
forecasts of its own plans were also less accurate. The leading models forecast the abandonment rate
of their own plans to within 0.3 percentage points of the simulated rate, whereas Haiku's forecasts differed
by 2.2 percentage points.

These analyses concern the simulated banks. In the next section, we test whether the models' forecast
ranking carries over to real organizational changes.

\section{Forecast Rankings on Real Organizational Changes}
\label{sec:external}

In this section, we address the second question in Section~\ref{sec:intro}: does the forecast ranking from
the simulated banks carry over to real organizational changes? We describe the comparison
(Section~\ref{sec:comparison}), how the real changes were selected and disguised
(Section~\ref{sec:recognition}), and the order of analysis decisions
(Section~\ref{sec:order}). We then report the agreement between the two rankings
(Section~\ref{sec:agreement}) and examine what drives it (Section~\ref{sec:uncertainty}).

\subsection{The Comparison}
\label{sec:comparison}

We rank the same models in two ways. The \emph{benchmark ranking} orders models by their forecast
improvement on the assigned interventions (Section~\ref{sec:scoring}), with ties in rounded values broken by the unrounded scores (e.g.,
74.2\% for Sonnet versus 73.7\% for Sol). For the \emph{real-change ranking},
we give each model a set of real organizational changes, using only information available before each
change, and ask it to forecast the effect of the change with a point estimate and an 80\% prediction
interval. Each change has one or more outcome cells, such as the average wait in the first three months, and the target in each is the change between the before and after windows, or the treated-minus-control difference for a randomized trial. We then score each forecast against the observed outcome. A model's real-change score is
\begin{equation}
  H = \frac{1}{E} \sum_{e=1}^{E} \big( S_{e} - S^{\mathrm{cal}}_{e} \big),
  \label{eq:realscore}
\end{equation}
where $E$ denotes the number of real changes for which the model remained eligible, $S_{e}$ represents
the model's scaled interval score on change $e$, averaged over its outcome cells, and $S^{\mathrm{cal}}_{e}$ corresponds to the score of a calibrated no-change forecast, a
forecast of zero effect with an interval width set from the variation observed before the change. Each interval score is divided by the standard deviation of the same before-to-after change measured at placebo dates before the real change (for trials, across random splits of the control arm), and the calibrated forecast's interval spans the 10th to 90th percentiles of those placebo changes. Lower scores are better, and
a negative score means the model outperformed the calibrated no-change forecast. Unlike the ratio used for
forecast improvement (Eq.~\eqref{eq:fi}), Eq.~\eqref{eq:realscore} uses a difference, because
models are scored on different sets of eligible changes after the recognition checks, and subtracting each change's calibrated no-change score keeps their scores comparable.

The comparison covers 12 models: the nine ranked models from Section~\ref{sec:results}, together with
Muse Spark~1.3, Qwen3.8~27B and Gemini~3.5 Flash-Lite, which made one benchmark run per bank. The fourth
additional model, Solar Pro~4, did not meet the ranking criterion in Section~\ref{sec:setup} and is not
included. We measure agreement between the two rankings with Spearman's rank
correlation, $\rho$, which equals 1 when the rankings are identical and 0 when they are unrelated.

\subsection{Real Changes and Recognition Checks}
\label{sec:recognition}

A retrospective test is only informative if the models do not already know the outcomes. We therefore
disguised every real change and checked whether models recognized it.

We began with 38 candidate changes, of which a first recognition screen retained ten. We disguised the
other changes more strongly and re-admitted them, replacing the change-level screen with a check of each
model's knowledge of each outcome. The final set has 33 changes: the ten retained changes and 23 of the
re-disguised changes.

\paragraph{Disguising.}
For the ten retained changes, we replaced
organization and unit names, shifted dates by whole years chosen so
that weekdays were unchanged, and replaced identifiers with hashed codes. For the remaining 23, which used the revised procedure, we described
time relative to the change, used generic organization types, and described how each intervention worked
rather than naming the product. Within each source, we also rescaled all counts and all durations by hidden random
factors.\footnote{Counts were multiplied by a factor drawn from $[0.5, 2.0]$ and durations by a factor
drawn from $[0.7, 1.4]$, separately for each source.} This hides absolute values while preserving rates
such as the share of abandoned calls.

\paragraph{Recognition checks.}
In an initial probe, models frequently identified the source of a change (at least one model flagged 37 of the 38 candidate changes). However,
naming a source does not imply knowing the outcome. A second, pre-registered probe therefore asked each
model about the outcomes directly. It also included placebo questions: either the same organization and type of change at a date two to three years away, when no change took place, or a plausible change that never happened. The prompt did not say which items were placebos. We set aside 29 model--change pairs for
which the probe indicated prior knowledge of the outcome, mostly for published trials. These pairs were not run, and each model is scored only on the changes for which it remained eligible. As a separate
sensitivity check, we also removed, for each model, every change whose source that model had identified.

\paragraph{Final set of changes.}
The final set contains 33 real changes from four sources: open-source software teams adopting AI coding
assistants or agents, government contact centers, randomized workplace trials, and the contact center of a
US bank (Bank~B).\footnote{The sources contribute 16, nine, six and two changes,
respectively.} A third of the changes concern contact centers.

\subsection{Order of Analysis Decisions}
\label{sec:order}

We fixed the test plan, including the agreement criteria in Table~\ref{tab:verdict}, on 29 October 2026,
before the benchmark runs. Before seeing any real-change scores, we also fixed the revised disguising
procedure described above. The 33-change set and the per-model recognition check (Table~\ref{tab:verdict},
marked $\dagger$) were specified on 2 November 2026, after the benchmark results were available.

\subsection{Agreement Between Rankings}
\label{sec:agreement}

The comparison met all five pre-specified criteria for agreement, as well as the recognition check
added later (Table~\ref{tab:verdict}). Across the 12 models, $\rho=0.92$. A 95\% percentile bootstrap interval, obtained from 2,000 resamples of
the real changes with randomized trials resampled by study, is 0.50--0.92. The upper bound coincides with
the observed value because $\rho$ is close to its maximum, so resampling can lower the agreement much more
than it can raise it. The interval holds the benchmark ranking fixed and therefore does not reflect
uncertainty in the benchmark scores. A one-sided permutation test of the rank correlation, which shuffles
the models' real-change scores 10,000 times, gave $p=0.0002$. This differs from the test in
Table~\ref{tab:verdict}, which asks whether the models' real-change scores differ from one another at all
($p=0.07$). That weaker result means that differences among models on the real changes are only modestly
resolved, which is why ranks in the middle of the table reorder easily.

\begin{table}[t]
\centering
\small
\caption{Criteria for agreement between the benchmark and real-change rankings (12 models, 33 changes).
The omission checks report the lowest correlation after removing each source or each model in turn.
$^\dagger$Specified after the benchmark results were available (Section~\ref{sec:order}).}
\label{tab:verdict}
\begin{tabular}{lll}
\toprule
Criterion & Threshold & Result \\
\midrule
Rank correlation ($\rho$) & $\geq 0.6$ & 0.92 \\
Lower bound of 95\% bootstrap interval & $> 0$ & 0.50 \\
Lowest $\rho$ after omitting one source & $\geq 0.4$ & 0.73 (US bank omitted) \\
Lowest $\rho$ after omitting one model & $\geq 0.4$ & 0.90 \\
Models differ on real changes (permutation test) & $p < 0.2$ & $p=0.07$ \\
$\rho$ excluding recognized changes$^\dagger$ & $\geq 0.4$ & 0.83 \\
\midrule
Verdict & & \textbf{agree} \\
\bottomrule
\end{tabular}
\end{table}

\begin{figure}[htbp]
\centering
\begin{minipage}[c]{0.6\linewidth}\centering
\resizebox{\linewidth}{!}{%
\begin{tikzpicture}[font=\scriptsize, y=-0.4cm, x=0.75cm]
  \node[font=\small\bfseries, anchor=west] at (-4.6,-1.2) {(a)};
  \node[font=\scriptsize\bfseries, text=rbInk, anchor=east] at (0.2,0) {Benchmark rank};
  \node[font=\scriptsize\bfseries, text=rbInk, anchor=west] at (3.8,0) {Real-change rank};
  \foreach \i/\name in {1/Astra, 2/Sonnet, 3/Sol, 4/Fable, 5/Muse Spark 1.3,
    6/Haiku, 7/DeepSeek, 8/Luna, 9/Qwen Flash, 10/Gemini Flash, 11/Qwen3.8 27B,
    12/Gemini 3.5 Flash-Lite} {
    \node[anchor=east, text=rbInk] at (-0.15,\i+0.5) {\name~\i};
  }
  \foreach \j/\name in {1/Astra, 2/Sol, 3/Muse Spark 1.3, 4/Fable, 5/Sonnet,
    6/Haiku, 7/Qwen Flash, 8/DeepSeek, 9/Luna, 10/Qwen3.8 27B, 11/Gemini Flash,
    12/Gemini 3.5 Flash-Lite} {
    \node[anchor=west, text=rbInk] at (4.15,\j+0.5) {\j~\name};
  }
  \foreach \a/\b in {1/1, 2/5, 3/2, 4/4, 5/3, 6/6, 7/8, 8/9, 9/7, 10/11, 11/10, 12/12} {
    \draw[rbGray!90!black, line width=0.9pt] (0,\a+0.5) -- (4,\b+0.5);
    \fill[rbGrayL] (0,\a+0.5) circle (1.9pt); \draw[rbInk] (0,\a+0.5) circle (1.9pt);
    \fill[rbGrayL] (4,\b+0.5) circle (1.9pt); \draw[rbInk] (4,\b+0.5) circle (1.9pt);
  }
\end{tikzpicture}}
\end{minipage}%
\begin{minipage}[c]{0.37\linewidth}\centering
\resizebox{\linewidth}{!}{%
\begin{tikzpicture}

\begin{axis}[
  width=4.5cm, title={(b)}, title style={font=\small\bfseries, at={(-0.35,1.0)}, anchor=south west}, height=4.6cm, clip=false, xbar, bar width=8pt, xmin=0, xmax=112, xtick={0,25,50,75,100},
  symbolic y coords={LMArena,Output price,RehearsalBench}, ytick=data,
  yticklabels={LMArena$^{*}$,Output price,RehearsalBench}, enlarge y limits=0.3,
  xlabel={\% of model pairs ordered correctly}, tick label style={font=\scriptsize, text=rbInk},
  label style={font=\scriptsize, text=rbInk}, axis line style={rbInk!60}, tick style={rbInk!60},
  xmajorgrids, grid style={rbGrayL},
  nodes near coords, nodes near coords style={font=\scriptsize, text=rbInk},
  every node near coord/.append style={anchor=west}]
\addplot[fill=rbTealL, draw=rbTeal] coordinates {(60,LMArena) (67,Output price) (89,RehearsalBench)};
\draw[rbRedS, dashed, thick] ([yshift=-10pt]axis cs:50,LMArena) -- ([yshift=12pt]axis cs:50,RehearsalBench);
\node[font=\scriptsize, text=rbRedS!80!black, anchor=south] at ([yshift=12pt]axis cs:50,RehearsalBench) {chance};
\end{axis}
\end{tikzpicture}}
\end{minipage}
\caption{Agreement with real organizational changes. (a) Forecast ranks on the simulated banks and on the
33 real changes. (b) Share of model pairs ordered as on real changes, for the benchmark ranking, a ranking
by output price and the LMArena leaderboard of 27~October 2026 (66 pairs; 55 for LMArena, which does not
list every model). $^{*}$Post hoc (Section~\ref{sec:agreement}).}
\label{fig:ranks}
\end{figure}
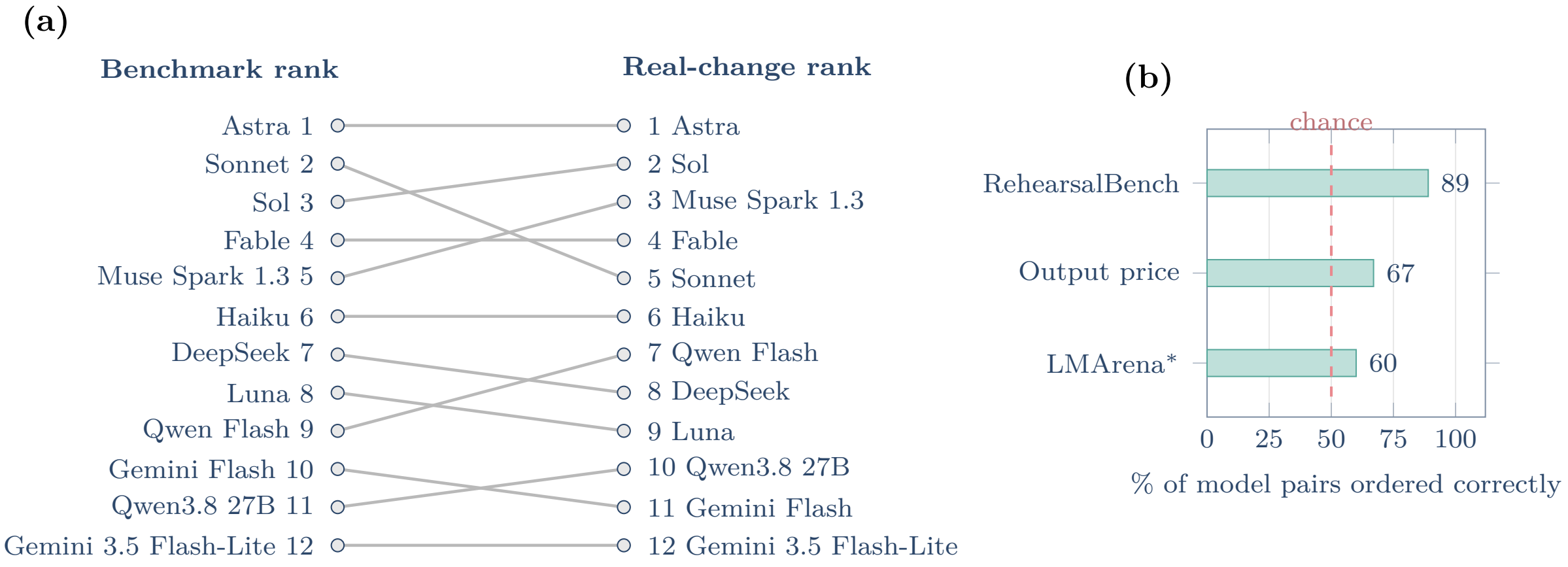

Figure~\ref{fig:ranks}a illustrates the two rankings. The same five models led both rankings, and the same
three trailed. Most models moved by at most one rank. The largest shift was Sonnet, which ranked second on
the benchmark but fifth on real changes: the real-change scores of the models ranked fourth to eighth lie within 0.2 of one another ($-1.18$ to $-0.98$), so small differences reorder them. The agreement is most
sensitive to the two US bank changes: omitting them lowers $\rho$ to 0.73, which still meets the criterion.

The benchmark ranking also ordered model pairs more accurately than two common alternatives
(Figure~\ref{fig:ranks}b). It identified the better forecaster on real changes in 89\% of model pairs. In
contrast, ranking models by their price per output token did so in 67\% of pairs, and the LMArena text
leaderboard, which ranks models by crowd-sourced pairwise preferences~\citep{chiang2024chatbot}, did so in
60\% of the 55 pairs among the 11 models it lists. This LMArena comparison is post hoc: our
pre-specified plan included LMArena only if it listed every model, and it does not list Qwen Flash.

\subsection{The Role of Uncertainty Estimates}
\label{sec:uncertainty}

Both evaluations reward intervals that are narrow enough to be informative yet wide enough to contain the
outcome. The agreement could therefore reflect how well models judge their uncertainty rather than how
accurately they estimate effect sizes. Consistent with this, interval coverage was strongly correlated
across the two settings ($\rho=0.78$).

This suggests that interval calibration contributes to the agreement. How much of
it reflects accurate estimates of effect size remains unclear. A high benchmark rank may therefore
reflect well-judged uncertainty as much as accurate point forecasts, and should be read accordingly.

\section{Conclusion and Future Work}
\label{sec:conclusion}

In this paper, we proposed \RB{}, a benchmark that evaluates how LLMs choose and forecast operational
changes in simulated bank contact centers. The simulator was built and validated on historical call records
and combined with documented assumptions about the effects of AI interventions. Most models chose plans
worth 5--7\% of the no-change cost, and every model predicted the direction of each effect correctly on average. Nevertheless,
models from all developers overstated the size of several effects, most notably the increase in repeat
contacts under a voice agent. A manager deciding whether to adopt such a technology needs to know not only
the direction of a change but also whether its size justifies the cost. Furthermore, the benchmark's
forecast ranking agreed closely with the models' ranking on 33 real organizational changes, although part
of this agreement may reflect how well models judge their own uncertainty.

Two design choices make these measurements possible. Because every plan is replayed on the same call sequences,
decision value isolates the effect of each plan from random variation. Because every model forecasts the
same assigned interventions, forecasts can be compared fairly even when models recommend different plans.
Our findings suggest that simulated rehearsal is a promising way to screen AI decision-support tools before
deployment. Candidate applications include contact-center management, workforce planning and other service
operations that can be modeled as queues.

This study focuses on bank contact centers, and its scope suggests several directions for future research:
\begin{enumerate}
\item \emph{Other domains:} The benchmark should be extended beyond contact centers, starting with the
  test on software teams adopting AI tools that we have designed.
\item \emph{Recent and observed data:} The simulator is calibrated on call records from 1999--2003, and
  the effects of AI interventions rest on documented ranges. Future studies should calibrate it on recent
  records and on observed outcomes of AI deployments as these become available.
\item \emph{A wider mix of interventions:} Staffing changes dominated the best-known plans in our banks.
  Future versions of the benchmark should include banks in which other interventions are the best choice, to test models'
  judgment about those interventions more fully.
\item \emph{Value in operating centers:} Our comparison tests whether forecast rankings carry over to real
  changes. Establishing the value of recommended plans will require evaluating them directly in operating
  centers.
\item \emph{Sources of error and training:} Future research is suggested to test whether the shared
  overstatements arise from the vendor materials given to the models, and to use the simulator as an
  environment for training agents with reinforcement learning.
\end{enumerate}

\section*{Reproducibility Statement}

We publish model-level results, the leaderboard and the figures at
\url{https://github.com/afterflowinc/rehearsalbench}. To protect future evaluations, we keep the
simulation engine, evaluator, scoring code, generated banks and real-change materials private. The call records
are available from the Technion SEELab archive.

\section*{Ethics and Disclosures}

The archive records are anonymized. Aggregate figures from a private contemporary contact center informed only the direction of some
parameter ranges, and we withhold the center's identity and figures. Claude Opus~5.5 was used as a coding
assistant in developing the benchmark software and was not evaluated (Section~\ref{sec:scoring}).

\setlength{\bibsep}{3pt plus 1pt minus 1pt}
\bibliographystyle{plainnat}
\bibliography{references}

\end{document}